\documentclass[sigconf]{acmart}
\AtBeginDocument{%
  }

\copyrightyear{2025}
\acmYear{2025}
\setcopyright{acmlicensed}\acmConference[MM '25]{Proceedings of the 33rd ACM International Conference on Multimedia}{October 27--31, 2025}{Dublin, Ireland}
\acmBooktitle{Proceedings of the 33rd ACM International Conference on Multimedia (MM '25), October 27--31, 2025, Dublin, Ireland}
\acmDOI{10.1145/3746027.3758244}
\acmISBN{979-8-4007-2035-2/2025/10}

\usepackage[utf8]{inputenc}

\usepackage{url}            % simple URL typesetting
\usepackage{booktabs}       % professional-quality tables
\usepackage{amsfonts}       % blackboard math symbols
\usepackage{nicefrac}       % compact symbols for 1/2, etc.
\usepackage{xcolor}         % colors

\usepackage{tabularx}
\usepackage{xspace}
\usepackage{multicol}
\usepackage{multirow}
\usepackage{tabularx}
\usepackage{amsmath}
\usepackage{bm}
\usepackage{textcomp}
\usepackage{graphicx}
\usepackage{caption}
\usepackage{cleveref}
\usepackage{subcaption}
\usepackage{placeins}
\usepackage{array}
\usepackage{tikz}
\usepackage{adjustbox}
\usepackage{footmisc}

\usetikzlibrary{positioning}

\newcommand{\textbra}{\textup{$\langle$}}
\newcommand{\textket}{\textup{$\rangle$}\xspace}

\usepackage[normalem]{ulem}
\useunder{\uline}{\ul}{}

\usepackage{wrapfig,lipsum,booktabs}

\newcolumntype{P}[1]{>{\centering\arraybackslash}p{#1}}
\newcolumntype{Y}{>{\centering\arraybackslash}X}
\newcolumntype{R}{>{\raggedleft\arraybackslash}X}
\newcommand{\blue}[1]{\textcolor{blue}{#1}}

\newcommand{\methodName}{LEGO\xspace}
\newcommand{\dataset}{TripAlign\xspace}
\newcommand{\methodFullName}{\textbf{L}arge 3D Vision-Language Model Learning from \textbf{E}gocentric \textbf{G}rounded \textbf{O}bservations\xspace}

\usepackage{hyperref}

\renewcommand{\shortauthors}{Wentao Mo, Qingchao Chen, Yuxin Peng, Siyuan Huang, \& Yang Liu}

\begin{document}

\title{Advancing 3D Scene Understanding with MV-ScanQA Multi-View Reasoning Evaluation and TripAlign Pre-training Dataset}
\author{Wentao Mo}
\affiliation{%
  \institution{Wangxuan Institute of Computer Technology, Peking University}
  \city{Beijing}
  % \state{Beijing}
  \country{China}
}
\email{mowt@pku.edu.cn}

\author{Qingchao Chen}
\affiliation{%
  \institution{National Institute of Health Data Science, Peking University}
  \city{Beijing}
  % \state{Beijing}
  \country{China}
}
\email{qingchao.chen@pku.edu.cn}

\author{Yuxin Peng}
\affiliation{%
  \institution{Wangxuan Institute of Computer Technology, Peking University}
  \city{Beijing}
  % \state{Beijing}
  \country{China}
}
\email{pengyuxin@pku.edu.cn}
\author{Siyuan Huang}
\affiliation{%
  \institution{State Key Laboratory of General Artificial Intelligence, BIGAI}
  \city{Beijing}
  % \state{Beijing}
  \country{China}
}
\email{huangsiyuan@ucla.edu}

\author{Yang Liu}
\affiliation{%
%   \institution{Wangxuan Institute of Computer Technology, Peking University}
% \institution{State Key Laboratory of General Artificial Intelligence}
  \institution{Wangxuan Institute of Computer Technology, State Key Laboratory of General Artificial Intelligence, \\Peking University}
  \city{Beijing}
  % \state{Beijing}
  \country{China}
}
\email{yangliu@pku.edu.cn}
\authornote{Corresponding author}

\renewcommand{\shortauthors}{Wentao Mo, Qingchao Chen, Yuxin Peng, Siyuan Huang, \& Yang Liu}

\begin{abstract}
The advancement of 3D vision-language (3D VL) learning is hindered by several limitations in existing 3D VL datasets: they rarely necessitate reasoning beyond a close range of objects in single viewpoint, and annotations often link instructions to single objects, missing richer contextual alignments between multiple objects. This significantly curtails the development of models capable of deep, multi-view 3D scene understanding over distant objects. 
To address these challenges, we introduce MV-ScanQA, a novel 3D question answering dataset where 68\% of questions explicitly require integrating information from multiple views (compared to less than 7\% in existing datasets), thereby rigorously testing multi-view compositional reasoning. 
To facilitate the training of models for such demanding scenarios, we present \dataset dataset, a large-scale and low-cost 2D-3D-language pre-training corpus containing 1M \textbra 2D view, set of 3D objects, text\textket triplets that explicitly aligns groups of contextually related objects with text, providing richer, view-grounded multi-object multimodal alignment signals than previous single-object annotations.
We further develop \methodName, a baseline method for the multi-view reasoning challenge in MV-ScanQA, 
transferring knowledge from pre-trained 2D LVLMs to 3D domain with \dataset.
Empirically, \methodName pre-trained on \dataset achieves state-of-the-art performance not only on the proposed MV-ScanQA, but also on existing benchmarks for 3D dense captioning and question answering.
Datasets and code are available at \url{https://matthewdm0816.github.io/tripalign-mvscanqa}.
\end{abstract}



% %%
% %% Keywords. The author(s) should pick words that accurately describe
% %% the work being presented. Separate the keywords with commas.

\maketitle

\section{Introduction}
\label{sec:intro}

\begin{figure}[htbp]
    \centering
    \adjustbox{scale={0.9}{0.8}}{\includegraphics[width=\linewidth]{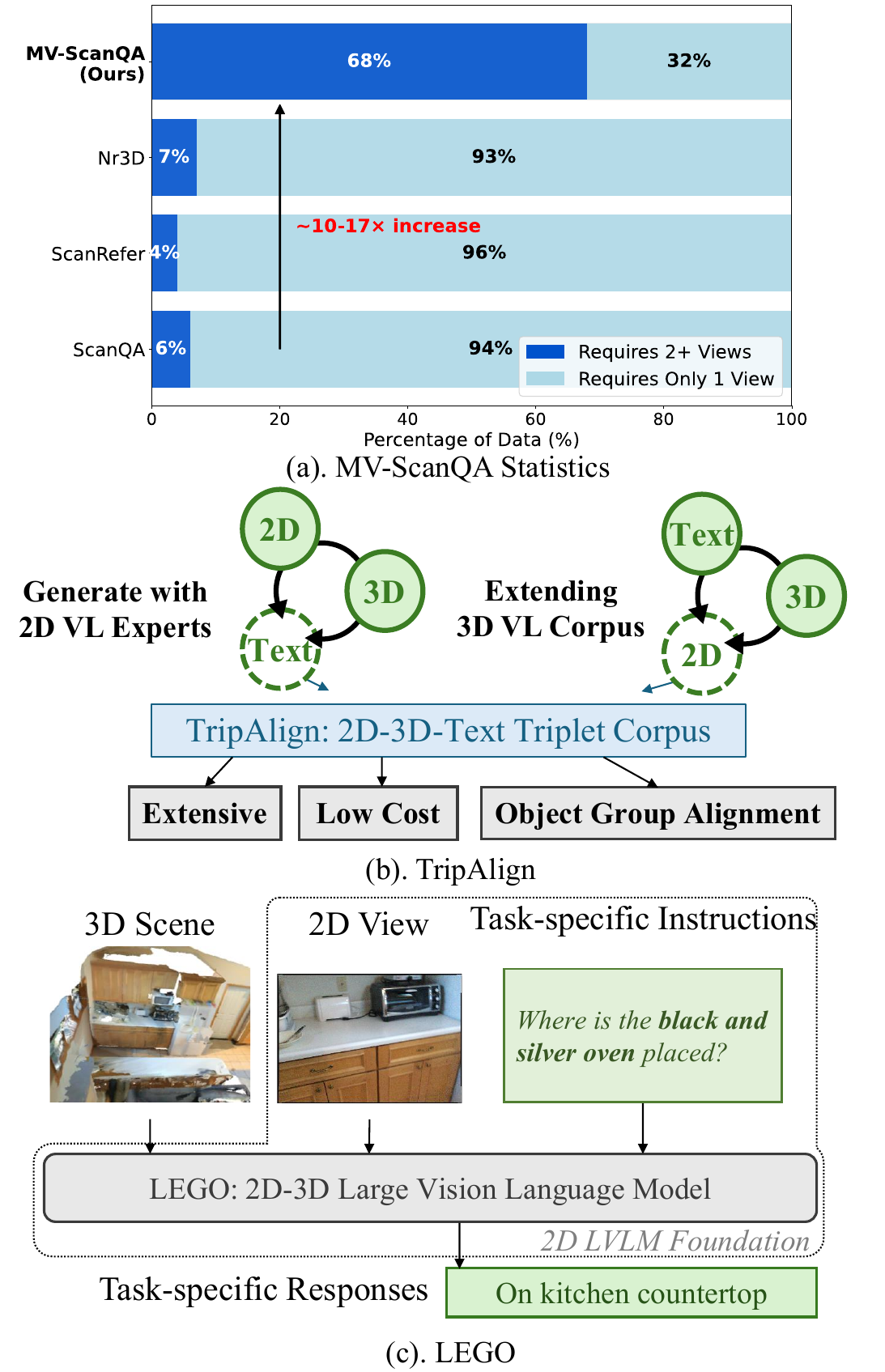}}
    \caption{
    (a). MV-ScanQA tests multi-view reasoning, with 10× more multi-view questions than existing datasets.
    (b).
    \dataset: A large-scale 2D-3D-language dataset that aligns object groups with captions for richer multi-modal supervision.
    (c). 
    \methodName: 
    Our 2D-3D LVLM leverages \dataset to transfer 2D LVLM knowledge to 3D tasks.
    }
    \label{fig:teasor}
\end{figure}

Recent advancements in Large Language Models (LLMs) \cite{openai2023gpt4, touvron2023llama} and their visual integration have led to powerful 2D Large Vision-Language Models (2D LVLMs) \cite{li2023blip, liu2023improved, liu2023visual}. However, extending these to 3D environments faces critical challenges stemming from  fundamental limitations in existing 3D vision-language datasets \cite{chen2021scan2cap, chen2020scanrefer, achlioptas2020referit_3d, azuma2022scanqa}:
\textbf{First, existing evaluation benchmarks \cite{chen2021scan2cap, chen2020scanrefer, achlioptas2020referit_3d, azuma2022scanqa} do not adequately test true 3D spatial understanding.} Our studies reveal that more than 93\% of tasks can be effectively solved\footnote{94\%, 96\% and 93\% for ScanQA, ScanRefer and Nr3D, respectively. Please refer to \Cref{sec:mvscanqa} for solvability explanation.} using information from just a single egocentric view. This reliance on single viewpoints means that models are not compelled to develop deeper, multi-view spatial reasoning capabilities essential for genuine 3D understanding. 
\textbf{Second, current training datasets suffer from sparse annotations.} Existing datasets \cite{chen2021scan2cap, chen2020scanrefer, achlioptas2020referit_3d, azuma2022scanqa} only link instructional text to single objects, missing the rich contextual relationships between multiple objects that naturally occur in scene descriptions, limiting models' ability to learn from dense and complex multi-object-to-text correspondences.

%%% MV-ScanQA
To address limited evaluation of multi-view 3D scene understanding, we introduce a new challenging \textbf{evaluation dataset MV-ScanQA}. MV-ScanQA is specifically designed to assess multi-view compositional reasoning by explicitly requiring information integration across multiple viewpoints. 
We define an instruction's "solvability" based on whether all relevant objects are sufficiently visible within a set of "solvable views". In stark contrast to existing datasets where less than 7\% of questions require multiple views, 68\% of questions in MV-ScanQA necessitate at least two distinct views to answer correctly, and 13\% require three or more views. This design pushes models beyond single-view comprehension, creating a substantially more challenging benchmark for true 3D scene understanding.

%%% Pretrain/TripAlign
To address the sparse annotation problem, we introduce \textbf{pre-training dataset \dataset}, a large-scale 2D-3D-language corpus with over 1M triplets. Unlike previous datasets that only provide single object-text pairs, \dataset achieves denser multi-object annotations by leveraging 2D views as an intermediary. Specifically, we form \textbra 2D view, set of 3D objects, instruction text\textket triplets, where the 2D view serves as a bridge to naturally group multiple contextually related objects that appear together. This approach enables us to automatically extract rich multi-object relationships from scene descriptions, providing richer cross-modal alignment signals for training. 
\dataset is constructed via a fully-automated, dual pipeline: (1) generating instruction text from the original 2D views of the 3D scenes 
by image captioners, naturally capturing multi-object relationships visible in the view, and (2) extending existing 3D VL datasets by selecting informative views that contain multiple relevant objects.

%%% LEGO
Finally, to demonstrate the utility of our proposed datasets, we present \textbf{\methodName} (\methodFullName) as a baseline solution. \methodName is a novel 2D-3D LVLM that uniquely builds upon pre-trained 2D LVLMs and effectively transfers their knowledge to the 3D domain. It leverages the instruction-relative egocentric views and the view-dependent multi-object alignment mechanism trained on \dataset  to process information from multiple viewpoints, showcasing strong performance on MV-ScanQA and other established 3D vision-language tasks.

We make the following contributions:
1) \textbf{MV-ScanQA}: A novel and challenging 3D question answering benchmark designed to evaluate multi-view scene understanding and compositional reasoning capabilities, where 68\% of questions require integrating information from multiple views.
2) \textbf{In-depth Solvability Analysis}: A quantitative analysis demonstrating the limitations of existing benchmarks in testing multi-view reasoning and highlighting the significant step forward provided by MV-ScanQA.
3) \textbf{\dataset}: A large-scale 2D-3D-language tri-modal aligned corpus of over 1M triplets that explicitly aligns object groups and egocentric views with instruction texts, addressing data scarcity and providing richer view-centric multimodal correspondence signals than previous single-object annotations.
4) \textbf{\methodName}: A 2D-3D LVLM that, when trained on our TripAlign dataset, achieves state-of-the-art performance on existing 3D VL benchmarks and on our new MV-ScanQA evaluation set. Its success highlights both the critical role of high-quality multi-view training data and the limitations of current 3D LVLMs in multi-view reasoning, validating the necessity of \dataset and MV-ScanQA in advancing the 3D VL field.

\section{Related Work}

\textbf{3D Vision-Language (3D VL) Datasets.} 
% There are emerging datasets enabling diverse 3D scene understanding \cite{zhu20233dvista, jia2024sceneverse, 3DSyn} and generation \cite{10.1145/3664647.3681653, Qin_2025_CVPR}. 
Existing \textbf{3D VL evaluation datasets} \cite{chen2020scanrefer, azuma2022scanqa, achlioptas2020referit_3d} primarily test single-view understanding, with over 93\% instructions solvable from one viewpoint. This fails to evaluate true 3D spatial reasoning capabilities. MV-ScanQA addresses this gap by requiring enforced multi-view integration, with 68\% of questions needing at least two views, creating a substantially more challenging benchmark that pushes models beyond single-view comprehension.
Existing \textbf{3D VL pre-training datasets} focus on combining data across multiple tasks to improve 3D VLMs' understanding abilities. Previous work \cite{zheng2023learning, zhu20233dvista, jia2024sceneverse, huang2023embodied, 3DSyn} collected diverse 3D VL datasets and generated captions from scene graphs. However, these datasets share a critical limitation: they only align instructions with \textit{single object}, missing rich multi-object-to-text alignments for scene understanding. Recently, SceneLLM and SceneVerse \cite{fu2024scenellm, jia2024sceneverse} generate data from 2D views, but still they overlook the detailed visual semantic correspondence between 2D views and 3D/language modality.
Distinct from these works that still maintain single-object focus, \dataset pioneers \textit{multi-object alignment} by explicitly linking groups of contextually related objects with instructions to offer denser multi-modal alignment signals. And distinct from datasets ignoring 2D correspondences, we are the first to construct true 2D-3D-text tri-modal triplets, enabling wide applications like knowledge transfer from 2D expert models.

\noindent\textbf{3D Vision-Language Pre-training (3D VLP) with Large Language Models (LLMs).} There are emerging methods for diverse 3D VL tasks like scene understanding \cite{zhou2025learn3dvqabetter, Ground_2025_ICRA, Schult23ICRA, chen2023vote2capdetr} and generation \cite{yi2023gaussiandreamer, 10646560, Qin_2025_CVPR, poole2023dreamfusion, Liang_2024_CVPR,10.1145/3664647.3681653}.
Recent 3D VLP works have sought to integrate LLMs \cite{openai2023gpt4, touvron2023llama, jiang2023mistral, jiang2024mixtral} with 3D vision \cite{3dllm, chen2023ll3da, yin2023lamm, huang2023embodied, zheng2023learning}, building upon current 3D scene representations.
However, these methods often rely on 3D object detectors, which can overlook important visual context like out-of-vocabulary objects. 
More recent approaches \cite{huang2024chat, fu2024scenellm} incorporate 2D views, but they typically process all views indiscriminately, and fail to leverage pretrained 2D LVLM capabilities. This overlooks the importance of selecting instruction-relevant views, leading to computational inefficiency and suboptimal performance.
In contrast, we advocate for building 3D LVLMs upon the foundation of 2D LVLMs, not just text-based LLMs. This approach leverages the 2D models' rich understanding of visual concepts. By training our model on our proposed 2D-3D-language data and filtering out irrelevant objects, we enable efficient knowledge 2D-3D knowledge transfer and more precise multi-modal alignment with less distracting objects.

\section{MV-ScanQA: A Multi-View 3D QA Dataset For Compositional Reasoning}
\label{sec:mvscanqa}

The development of 3D LVLMs capable of complex, multi-view 3D scene understanding in real scenarios is significantly hampered by the lack of benchmarks that genuinely probe multi-view spatial reasoning. To address this critical gap, we architected MV-ScanQA, a novel question-answering dataset specifically engineered to necessitate compositional reasoning across multiple visual perspectives.

\begin{table}[htbp]
\caption{Exemplar question composition in MV-ScanQA. }
\centering
\fontsize{8.5}{10}\selectfont
\begin{tabularx}{\linewidth}{p{5.5cm}R}
\toprule
Question & Answer \\ \midrule
Q1: On which side of the brown wooden \textbf{desk} is the \textbf{waste basket} located?  & \blue{on right side} \\ \midrule
Q2: What is the wooden \textbf{chair} in front of?  & \blue{small \textbf{desk}} \\ \midrule \midrule
Combined: What is on the right of the small \underline{\textbf{desk} where the wooden \textbf{chair} is in front of}? & \blue{\textbf{waste basket}} \\ \bottomrule
\end{tabularx}
\label{tab:example-mvscanqa}
\end{table}

\begin{figure}[htbp]
    \centering
    \includegraphics[width=\linewidth]{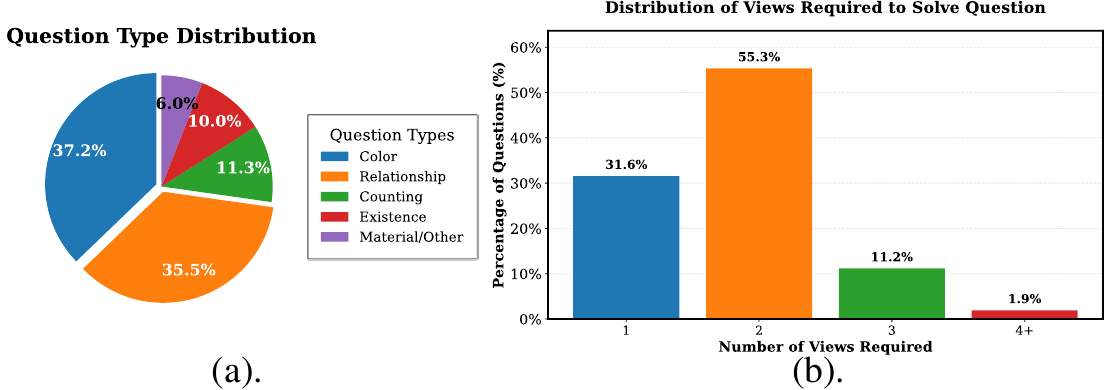}
    \caption{
    MV-ScanQA evaluation dataset analyses.
    (a). Question type distribution.
    (b). View-requirement analysis.
    }
    \label{fig:mv-scanqa-stats}
\end{figure}

\begin{figure*}[htbp]
    \centering
    \adjustbox{scale={0.9}{0.75}}{\includegraphics[width=\linewidth]{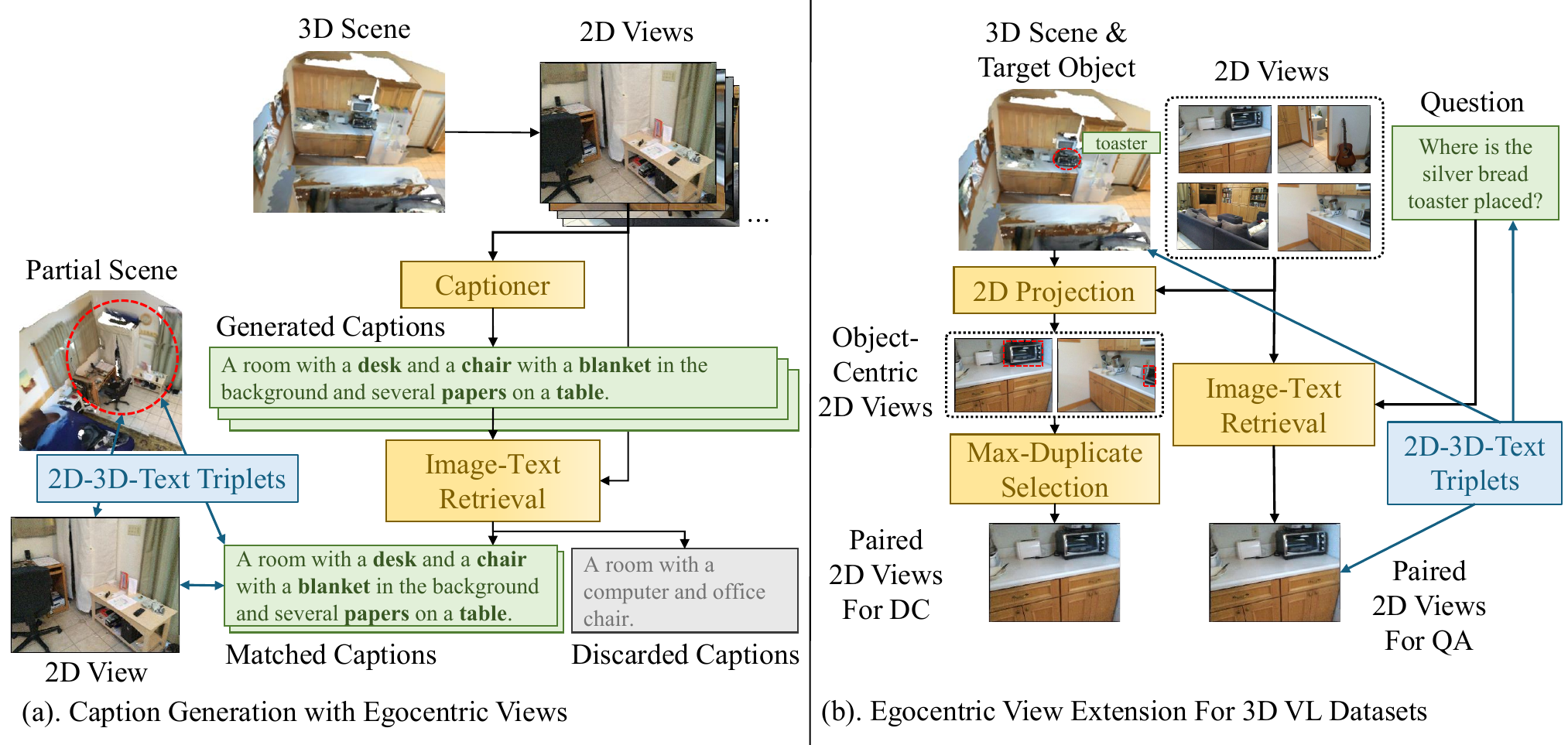}}
    \caption{
    Constructing \textbf{\dataset}:
    (a). 
    We sample 2D views described by an image captioner, and filtered by an image-text retrieval model.
    (b). 
    Without accessing task output, we select informative views for existing 3D VL datasets.
    }
    \label{fig:framecap}
\end{figure*}

\subsection{The "Single-View Bottleneck": Quantifying the Limits of Existing Benchmarks}
\label{subsec:solvability}
Before constructing MV-ScanQA, we conducted a "solvability analysis" to rigorously quantify the limitations of the prevalent 3D vision-language datasets \cite{achlioptas2020referit_3d, chen2020scanrefer, azuma2022scanqa}. We define a 3D VL instruction as "solvable" under a given set of egocentric views if all its semantically relevant objects are "witnessed", i.e. sufficiently visible, by at least one view in that set. An object is considered "witnessed" if its 2D projection onto an egocentric view overlaps with the view's image area with an Intersection-over-Smaller-Area (IoSA) exceeding 0.5. The IoSA metric between two 2D area $A, B$, calculated as 
$IoSA(A, B)=\frac{Area(A \cap B)}{\min (Area(A), Area(B))}\label{eqn:iosa}$,
robustly handles partially visible objects while disregarding those that are impractically small due to distance. Our analysis astonishingly revealed that in existing benchmarks like ScanQA \cite{azuma2022scanqa} 
, ScanRefer \cite{chen2020scanrefer}, and Nr3D \cite{achlioptas2020referit_3d}, only 6\%, 4\%, and 7\% of the instructions, respectively, require more than a single view to solve. This underscores a significant "\textbf{single-view bottleneck}", demonstrating that current evaluations largely fail to engage models in complex, multi-view spatial reasoning on long-range objects.

\subsection{Crafting Complexity in MV-ScanQA}
MV-ScanQA is designed to directly overcome this bottleneck. Instead of collecting new annotations from scratch, we developed a principled, LLM-guided framework for \textit{compositional question synthesis}. Conceptually, new questions are synthesized from a set of questions with shared object focus, as shown in \Cref{tab:example-mvscanqa}. This framework transforms existing single-view-solvable questions into more complex queries that demand genuine multi-view understanding and relational reasoning. The process involves two carefully designed stages:

\noindent \textbf{Question Candidates Pairing}
To ensure that newly synthesized questions achieve a higher order of complexity, we select candidate question pairs based on two crucial semantic criteria:
1) shared contextual anchor: the sets of objects relevant to each question in a pair must possess at least one common object. This common object acts as a semantic bridge, enabling meaningful composition.
2) complementary information requirement: neither question's set of related objects can be a complete subset of the other. This constraint guarantees that each original question contributes unique information, making it essential to consult both to answer the synthesized query.
Specifically, from all question pairs $q_1, q_2$ and their relative object anchor set $O_1, O_2$, we use them to synthesize new compositional question if $O_1 \cap O_2 \neq \emptyset$ and $O_1 \not\subseteq O_2$ and $O_2 \not\subseteq O_1$.
This selection procedure ensures that the combined question inherently requires a broader understanding of the scene's objects than either precursor question alone.

\noindent \textbf{Principled Question Composition}
Once suitable pairs are identified, we leverage the advanced generative capabilities of LLM (\verb|claude-3.5-sonnet|) to synthesize new questions, guided by two key directives:
1) \textbf{Integrative Complexity}: The LLM is instructed to weave together informational requirements from both parent questions into a single, novel query. This ensures the new question demands multi-faceted reasoning relative to the union of the object sets involved in its precursor questions.
2) \textbf{QA Verifiability}: The LLM is instructed to synthesize questions clearly stated and paired with an accurate, unambiguous, and definitive answer.
As shown in \Cref{tab:example-mvscanqa}, our composition process ensures that MV-ScanQA is populated with clear and verifiable questions demanding multi-view understanding and compositional reasoning.

\noindent\textbf{Solvablity Analysis of MV-ScanQA}
This multi-stage synthesis process results in approximately 10K new question-answer pairs for MV-ScanQA. As depicted in \Cref{fig:mv-scanqa-stats}(b), the efficacy of our approach is evident in the dataset's statistics: \textbf{68\% of the questions in MV-ScanQA require information from two or more distinct views to be answered correctly}, a stark contrast to the less than 7\% in the original ScanQA. Furthermore, 13\% of questions demand insights from three or more views. This distribution confirms that MV-ScanQA provides a significantly more challenging and realistic testbed for true 3D scene understanding and the compositional reasoning capabilities of advanced 3D LVLMs. 
Manual verification on 100 images shows 94\% questions are synthesized correctly.

\noindent\textbf{Splits and Metric} 
We follow common practices \cite{azuma2022scanqa, ma2022sqa3d, chen2020scanrefer} to split train/val/test splits based on the underlying 3D scene's split in ScanNet \cite{dai2017scannet}, and evaluate the answers using EM (\textbf{E}xact \textbf{M}atch), the same as previous QA benchmark \cite{azuma2022scanqa}.

\section{\dataset: A Pre-training Dataset with Multi-Object Alignment}
\label{sec:dataset}

While MV-ScanQA poses a new challenge on multi-view 3D scene understanding, training models to excel at such tasks requires addressing the fundamental limitation of sparse single-object annotations in existing 3D datasets.
To overcome this sparsity and provide denser supervision signals, we introduce \dataset, a large-scale 2D-3D-language corpus comprising over 1M \textbra 2D view, set of 3D objects, text\textket triplets, using 2D views as an intermediary to naturally group multiple contextually related objects that appear together, enabling automatic extraction of rich multi-object relationships without manual annotation effort. 
This approach transforms the sparse single-object paradigm into denser multi-object alignment annotations, where each triplet captures complex object interactions visible from a particular viewpoint which could facilitate 2D-to-3D knowledge transfer.

\dataset is built with two automated strategies (\Cref{fig:framecap}) designed to maximize both scale and quality:
1) \noindent \textbf{Generating Captions from Egocentric Views:} 
As shown in \Cref{fig:framecap}(a), this strategy creates a large collection of 2D view-text pairs grounded in 3D scenes. 
We begin with numerous egocentric views (e.g., ScanNet \cite{dai2017scannet}) from datasets captured by human annotaters walking in rooms. 
For each view, a pre-trained captioner \cite{liu2023improved, openai2023gpt4} generates multiple descriptions. To ensure high quality, an image-text retrieval model \cite{radford2021learning} filters these captions, retaining only the pairs with a strong semantic match. The result is a high-fidelity triplet: \textbra 2D view, visible 3D objects, textual caption\textket, which connects rich descriptions to specific groups of objects from a distinct viewpoint, rather than to a single object or the entire undifferentiated scene.
2) \noindent \textbf{Contextual View Extension for Existing 3D VL Datasets:} 
As shown in \Cref{fig:framecap}(b), this strategy augments established 3D vision-language datasets (for 3D question answering (QA), dense captioning (DC), etc.) by adding relevant 2D visual context. For each instruction in a 3D dataset, we select the most informative egocentric view from the scene.
For QA tasks, we use a retrieval model \cite{li2022blip} to pick the view most semantically similar to the question text.
For DC tasks, we choose the view that best captures the target object using IoSA in \ref{eqn:iosa} as a visibility metric. 
These chosen most relevant 2D views enrich current 2D VL datasets with detailed visual attributes and textures often lost in pure 3D representations.

\section{\methodName: Adapting 2D LVLM for 3D Scene Understanding}
\label{sec:method}

\begin{figure*}[htbp]
    \centering
    \adjustbox{scale={0.9}{0.75}}{\includegraphics[width=\linewidth]{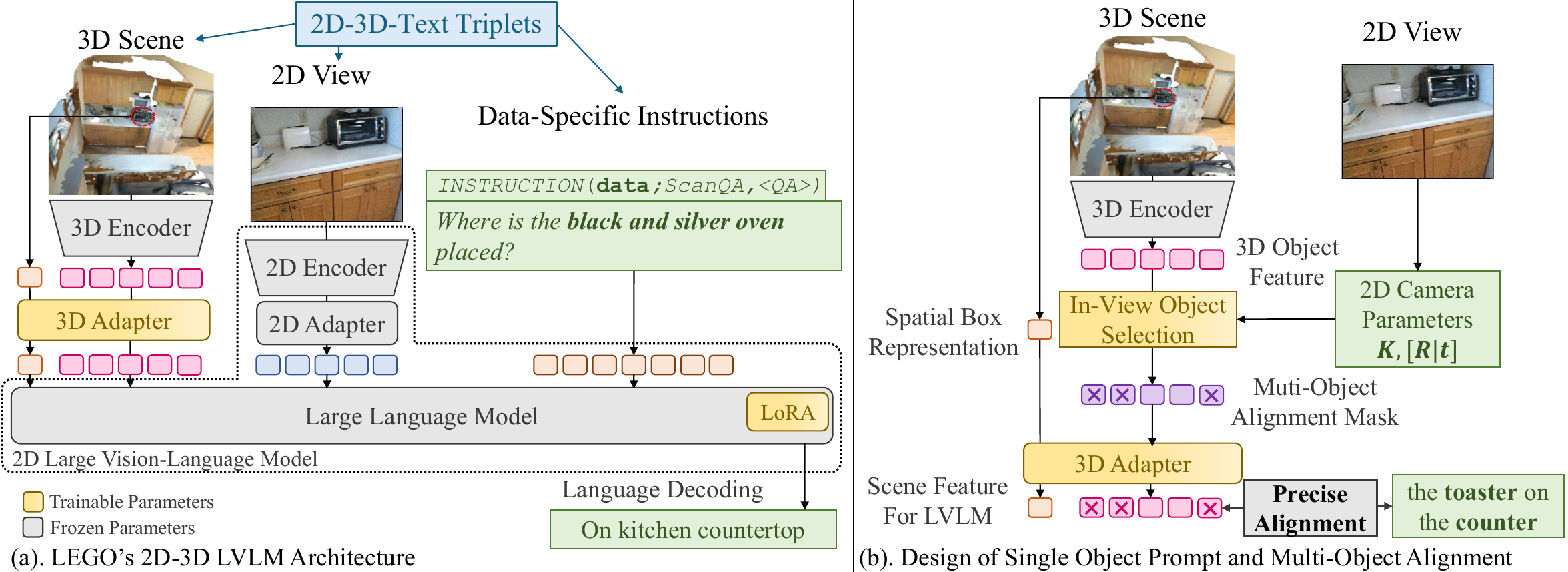}}
    \caption{
    \textbf{\methodName{} Overview}.
    (a). 
    \textbf{\methodName{}} builds upon a 2D LVLM and accepts 3D scene input. 
    (b). 
    Our proposed \textit{View-Dependent Multi-Object Alignment} mechanism removes irrelative objects with text, offering compact multi-modal alignment.
    }
    \label{fig:kuri}
\end{figure*}

To demonstrate the utility of \dataset and to provide a strong baseline for the 3D  VL tasks, particularly MV-ScanQA, we present \methodName, a 3D Large Vision-Language Model (LVLM) architected to effectively transfer knowledge from pre-trained 2D LVLMs to the 3D domain. As depicted in \Cref{fig:kuri}(a), \methodName takes as input a 3D scene, a text instruction, and informative 2D egocentric views\footref{fn:1}. These key views are chosen using the selection strategy from \Cref{sec:dataset} to ensure they are potentially "solvable" for the instruction.

\noindent\textbf{Multi-Modal Encoding.}
We build \methodName upon Fuyu \cite{fuyu-8b} 2D LVLM to leverage the established alignment of visual concept and language and transfer it to the 3D domain. 
As depicted in \Cref{fig:kuri}, a pre-trained 3D object detector $\mathcal{E}^{3D}$ \cite{chen2023end} extracts $M$ object features $o_{1:M} = \mathcal{E}^{3D}(\bm P)\in \mathbb{R}^{M\times d_{3D}}$ from the scene's point cloud $\bm P$.
In parallel, the LVLM's native image encoder $\mathcal{E}^{2D}$ extracts $N$ image patch features $p_{1:N}$ from the selected 2D view$\mathcal{I}$\footnote{2D views are selected without any access to task response.\label{fn:1}}.
Both sets of features are then projected, concatenated, and fed into the language model as a combined multi-modal prefix: $h_{\text{mm}} = [\operatorname{MLP}_{2D}(p_{1:N});\operatorname{MLP}_{3D}(o_{1:M})] \in \mathbb{R}^{(N+M)\times d}$.

\noindent\textbf{Precise 2D-3D Alignment with Leveraging Egocentric Views}
When processing the \textbra 2D view, set of 3D objects, text\textket triplets from \dataset where text is generated exclusively from an egocentric view, we propose a \textbf{view-dependent multi-object alignment mechanism} as depicted in \Cref{fig:kuri}(b), utilizing the unique granularity of our proposed \dataset dataset that only aligns the visible part of 3D scene in egocentric view with other modalities. 
Specifically, for the $M$ detected object proposals $\{o_i\}$, it filters them such that only objects clearly visible in the selected 2D view $\mathcal{I}$ are retained: $\{o_i\} \leftarrow \{o_i | \forall i\in \{1,2,...,M\}, \operatorname{IoSA}(\text{box}_i^{2D}, \mathcal I) > \tau\}$, where $\text{box}_i^{2D}$ is the 2D projection of the $i$-th object's 3D box, calculated with the camera's intrinsic ($\bm{K}$) and extrinsic ($[\bm{R}|\bm{t}]$) parameters. 
IoSA is defined in
\Cref{subsec:solvability} and $\tau$ is a visibility threshold.
In this way, only objects that can be clearly viewed in certain 2D perspective are kept and aligned with other modalities, while out-of-view objects are ignored, allowing a more precise multimodal alignment.

\noindent\textbf{Training} 
Trainable parameters $\theta$ of \methodName are optimized using a standard auto-regressive objective, predicting the response based on all multi-modal inputs:
$\mathcal{L}(\theta) = -\sum_{i=a}^{|w|} 
        \log p_{\theta} (w_i | h_{\text{mm}}; w_{1:i-1})$. 
Here $|w|$ is the length of textual tokens of instructional text and response, and $a$ is the starting position of the response. We adopt LoRA \cite{hu2022lora} for efficient training.

\section{Experiments}
\label{sec:exp}

\subsection{Evaluation and Implementation Details}

We evaluate \methodName{} on our newly proposed MV-ScanQA benchmark and compare it with previous methods to demonstrate the challenges of multi-view reasoning. Additionally, following prior work \cite{chen2023ll3da, huang2023embodied, zhu20233dvista}, we evaluate on established 3D vision-language datasets: Dense Captioning (DC) tasks on ScanRefer \cite{chen2020scanrefer} and Nr3D \cite{achlioptas2020referit_3d}, and Question Answering (QA) tasks on ScanQA \cite{azuma2022scanqa}. 
For DC tasks, we report text similarity metrics as defined in Scan2Cap \cite{chen2021scan2cap}. 
For QA tasks, we report ``\textbf{E}xact \textbf{M}atch" (EM) as well as text similarity metrics as defined in ScanQA \cite{azuma2022scanqa}.

\begin{table}[thbp]
\caption{Performance comparison on MV-ScanQA. N represents the number of views sufficient to view all question-related objects. Best performance is in \textbf{bold}.}
\fontsize{8}{10}\selectfont
\begin{tabularx}{\linewidth}{cYYYYY}
\toprule
\multirow{2}{*}{Method} & \multicolumn{5}{c}{MV-ScanQA}       \\ \cmidrule{2-6} 
                        & All  & N=1  & N=2  & N=3  & N$\ge$4 \\ \midrule
BridgeQA \cite{Mo_Liu_2024}   &  24.6 & 24.4 & 24.6 & 24.9  &  25.6   \\
LL3DA \cite{chen2023ll3da}    & 19.6 &  21.3 & 17.8 & 22.9 &  23.3   \\
LEO \cite{huang2023embodied}   &  24.2  & 25.6 & 23.1  & 27.3  & 20.9   \\
ChatScene  \cite{huang2024chat}   &  23.3 &  23.4 & 23.0 & 24.5 &  20.9   \\ \midrule
LEGO (Single-View)      & 30.0 & 32.7 & 29.6 & 26.5 & 23.3    \\
LEGO (Multi-View)       & \textbf{34.1} & \textbf{35.1} & \textbf{34.3} & \textbf{30.9} & \textbf{30.2}    \\ \bottomrule
\end{tabularx}
\label{tab:mv-scanqa}
\end{table}

\begin{table}[thbp]
    \caption{
    Dense captioning performance comparison.
    Best and second-best performance are in bold and underlined. 
    }
\fontsize{8}{10}\selectfont
\centering
\begin{tabularx}{\columnwidth}{c|YY|Y}
\toprule
\multirow{3}{*}{Method}                     & \multicolumn{2}{c|}{Scan2Cap}      & Scan2Cap         \\
                                            & \multicolumn{2}{c|}{(ScanRefer)}   & (Nr3D)           \\
                                            & C@0.25           & C@0.5           & C@0.5            \\ \midrule
Scan2Cap \cite{chen2021scan2cap}            & 56.82            & 39.08           & 27.47            \\
3D-VLP \cite{zhang2023visionlanguage}       & 70.73            & 54.94           & -                \\
3D-VisTA \cite{zhu20233dvista}              & -                & 61.60           & -                \\
Vote2Cap-DETR++ \cite{chen2023vote2capdetr} & \underline{76.36} & 67.58           & 47.08            \\
LL3DA \cite{chen2023ll3da}                  & 74.17            & 65.19           & \underline{51.18} \\
LEO \cite{huang2023embodied}                & -                & \underline{72.4} & -                \\
ChatScene \cite{huang2024chat}              & 81.94            & 77.19           & -                \\
\methodName{} (Ours)                        & \textbf{84.66}   & \textbf{78.56}  & \textbf{61.36}   \\ \bottomrule
\end{tabularx}
\label{tab:perf-dc}
\end{table}

\begin{table}[t]
\caption{
Question answering performance comparison.
Best and second-best performance are in bold and underlined.
}
\fontsize{8}{10}\selectfont
\centering
\begin{tabularx}{\columnwidth}{c|YYYYY}
\toprule
\multirow{3}{*}{Method} & \multicolumn{5}{c}{ScanQA} \\
 & Val & \multicolumn{2}{c}{Test w/ obj} & \multicolumn{2}{c}{Test w/o obj} \\
 & EM & EM & C & EM & C \\ 
\midrule
ScanQA \cite{azuma2022scanqa} & 21.1 & 23.5 & 67.3 & 20.9 & 60.2 \\
3D-VLP \cite{zhang2023visionlanguage} & 21.7 & 24.6 & 70.2 & 21.6 & 63.4 \\
3D-VisTA \cite{zhu20233dvista} & - & 27.0 & 76.6 & 23.0 & 62.6 \\
BridgeQA \cite{Mo_Liu_2024} & 27.0 & \underline{31.3} & 83.8 & \underline{30.8} & 79.3 \\
LL3DA \cite{chen2023ll3da} & - & - & 78.2 & - & 70.3 \\
NaviLLM \cite{zheng2023learning} & 20.5 & 19.1 & 69.6 & - & - \\
LEO \cite{huang2023embodied} & 24.5 & 21.6 & 70.7 & 17.3 & 58.1 \\
Scene-LLM \cite{fu2024scenellm} & \underline{27.2} & - & - & - & - \\
ChatScene \cite{huang2024chat} & 21.6 & 25.5 & \underline{94.0} & 22.4 & \underline{79.8} \\
\methodName{} (Ours) & \textbf{28.4} & \textbf{33.7} & \textbf{96.6} & \textbf{32.7} & \textbf{91.5} \\
\bottomrule
\end{tabularx}
\label{tab:perf-qa}
\end{table}

\subsection{\methodName{} on MV-ScanQA}
To illustrate the challenging nature of MV-ScanQA and the straightforward extensibility of \methodName{} to a multi-view scenario, we first evaluate \methodName{} using only a single egocentric view. Then, we extend \methodName{} in a simplistic fashion by splicing four egocentric views in a 2x2 grid as input, while filtering out overly similar views. 
We observe (\Cref{tab:mv-scanqa})  that the single-view variant of \methodName{} already achieved decent performances. By simply extending \methodName{} to multiple views, it achieves a significant performance boost, particularly for questions requiring more than 4 views (+6.9\%). This demonstrates the benefit of incorporating multiple views and \methodName{}'s extensibility to more complex 3D reasoning tasks. 
Existing 3D LVLMs, primarily trained on data where a single view often suffices, struggle with the increased complexity of MV-ScanQA, showing a potential gap in current methods. This underscores the need for multi-view data in 3D VLP research, with \methodName{} and MV-ScanQA serving as a first step towards addressing this challenge.

\begin{table}[t]
\centering
\fontsize{8}{10}\selectfont
\caption{Ablation of view-dependent multi-object alignment. 
}
\begin{tabularx}{\linewidth}{cYYY}
\toprule
\multirow{2}{*}{Method}    & ScanRefer      & Nr3D           & ScanQA            \\
                           & C@0.25         & C@0.5          & EM             \\
\midrule
w/o Multi-Object Alignment      &  81.04   & 60.93  &  27.91   \\
\midrule
Full (Ours) & \textbf{84.66} & \textbf{61.36} & \textbf{28.43} \\
\bottomrule
\end{tabularx}
\label{tab:align}
\end{table}

\subsection{\methodName{} on 3D Vision-Language Tasks}

We quantitatively assessed \methodName{}'s ability to interpret and reason across downstream 3D VL tasks. As shown in \Cref{tab:perf-dc} for DC and \Cref{tab:perf-qa} for QA, \methodName{} achieved state-of-the-art results on all 9 metrics.
Scene-LLM \cite{fu2024scenellm} proposed a similar view-based data generation pipeline, but they ignored the 2D-3D knowledge transfer and fine-grained visual context that egocentric views could furnish, therefore falls behind our method on most metrics.
Notably, \methodName{} excelled in CiDER metrics on all datasets, which measure free-form similarity between model responses and human annotations, with significant improvements of +2.7, +1.4, +10.7 on ScanRefer and Nr3D splits, and +2.6, +11.7 on ScanQA splits. 
These results validate the effectiveness of \dataset's dense multi-object annotations and tri-modal alignment strategy, demonstrating how richer supervision signals from grouped object-text pairs enhance 3D scene understanding capabilities.

\subsection{Ablation Studies}
\noindent\textbf{View-Dependent Multi-Object Alignment.}
We ablate the use of the multi-object alignment strategy that associates a precise set of objects with each caption. Our method yields consistent performance improvements (\Cref{tab:align}) across all datasets (+3.6 on ScanRefer, +0.4 on Nr3D and +0.5 on ScanQA) compared to methods without in-view object selection. These results demonstrate the value of fine-grained multi-modal alignment, which reduces distracting objects and filters spurious correspondences during training, enabling effective 2D-3D transfer.

\begin{table}
\centering
\fontsize{8}{10}\selectfont
\caption{Ablations of proposed 2D-3D-language alignment data paradigms with \dataset. 
}
\begin{tabularx}{\linewidth}{cYYY}
\toprule

\multirow{2}{*}{Data Paradigm}    & ScanRefer      & Nr3D           & ScanQA           \\
                           & C@0.25         & C@0.5          & EM                  \\ \midrule
from scratch & 74.14 & 53.01 & 25.13  \\ \midrule
+Egocentric View Extension &  78.50  & 57.29 & 27.22  \\
+\dataset Extension & \textbf{84.66} & \textbf{61.36} & \textbf{28.43}  \\
\bottomrule
\end{tabularx}
\label{tab:data}
\end{table}

\noindent\textbf{Use of \dataset.}
In \Cref{tab:data}, we investigate the effectiveness two data pipelines of of \dataset: extending existing 3D VL datasets with 2D views, and generate captions with 2D VL experts. 
By collecting pre-established 3D VL datasets and pairing instruction-relative egocentric views for each data sample, the performance consistently improves on all datasets (+4.4, +4.3, +2.1 on ScanRefer, Nr3D, and ScanQA scores, respectively).
Furthermore, by extending existing 3D VL data with 2D-3D-text triplets generated with egocentric views and 2D VL expert models, \methodName achieves best performance (+10.5, +8.4, +3.3 on ScanRefer, Nr3D and ScanQA scores, respectively), confirming that the our comprehensive egocentric view data pipeline could alleviate the data scarcity challenge.

\section{Conclusions}
\label{sec:concl}

In this paper, we address two fundamental limitations in 3D vision-language research through complementary contributions.
First, we introduce MV-ScanQA, a challenging evaluation dataset where 68\% of questions require multi-view reasoning, more than 10×
Second, to overcome the sparse single-object annotations that limit current 3D VL models, we present \dataset, a large-scale pre-training dataset that pioneers dense multi-object annotations. By using 2D views to naturally group contextually related objects, \dataset provides richer supervision signals essential for complex scene understanding.
Our baseline \methodName{}, trained on \dataset, achieves state-of-the-art performance on existing benchmarks and strong results on MV-ScanQA while showing deficiencies of current 3D LVLMs in multi-view reasoning, validating the critical importance of dense multi-object annotations and proper multi-view evaluation for advancing 3D vision-language understanding.

% \section*{}
\noindent\textbf{Acknowledgements.} This work was supported by the
grants from the National Natural Science Foundation of
China (62372014, 62525201, 62132001, 62432001), Beijing Nova Program and Beijing Natural Science Foundation (4252040, L247006).

% \appendix

% % \maketitlesupplementary
% % \clearpage
% % \setcounter{page}{1}
% \input{src/06_appendix}

\bibliographystyle{ACM-Reference-Format}
\balance
\bibliography{main.tex_used}

\end{document}